\documentclass[12pt]{article}
\usepackage{amsfonts,latexsym,fullpage,amsmath,latexsym,amssymb}

\newcommand{\Y}{{\cal Y}}
\newcommand{\cP}{{\bf P}}
\newcommand{\N}{\mathbb N}

\newcommand{\x}{{\bf x}}
\newcommand{\qed}[0]{\hfill $\Box$}

\newtheorem{theorem}{Theorem}

\newtheorem{definition}[theorem]{Definition}

\title{Required sample size for learning sparse\\ Bayesian
networks with many variables}

\date{April 26, 2002}

\author{ Pawe{\l}
Wocjan\thanks{e-mail: {\protect\tt
\{wocjan,janzing,eiss\_office\}@ira.uka.de}}, Dominik Janzing, and Thomas Beth\\ 
\small Institut f{\"u}r Algorithmen und Kognitive Systeme,
Universit{\"a}t Karlsruhe,\\[-1ex] \small Am Fasanengarten 5,
D-76\,131 Karlsruhe, Germany}
\begin{document}

\maketitle

\abstract{Learning joint probability distributions on $n$ random
variables requires exponential sample size in the generic case.  Here
we consider the case that a temporal (or causal) order of the
variables is known and that the (unknown) graph of causal dependencies
has bounded in-degree $\Delta$. Then the joint measure is uniquely
determined by the probabilities of all $(2\Delta+1)$-tuples.  Upper bounds on
the sample size required for estimating their probabilities can be
given in terms of the VC-dimension of the set of corresponding
cylinder sets. The sample size grows less than linearly with $n$.}

\section{Introduction}
Learning joint probability measures on a large set of variables is an
important task of statistics. One of the main motivations to estimate
joint probabilities is to study statistical dependencies and
independencies between the random variables \cite{Pearl:00}. In many
applications the goal is to obtain information on the underlying
causal structure that produces the statistical correlations.  However,
the problem of learning causal structure from statistical data is in
general a deep problem and cannot be solved by statistical
considerations alone \cite{Pearl:00,Glymour}.

Here we do not focus on the problem of uncovering the causal
structure, we rather address the problem of learning the probability
distribution on a large set of variables. In general, the sample size
required for estimating an unknown measure on the variables
$X_1,\dots,X_n$ grows exponentially with $n$. Assume for simplicity
that each $X_j$ is a discrete variable with $d$ possible values. Then
the probabilities of $d^n$ possible outcomes have to be estimated. The
sample size can be decreased considerably if prior knowledge on the
possible correlations is given. Consider for example the trivial case
when no statistical dependencies are possible at all, i.e.,
\[
P(x_1,x_2,\dots,x_n)=P(x_1)P(x_2)\dots P(x_n)\,,
\]
where $x_j$ denotes particular realizations of the corresponding
variable $X_j$. Then one has only to learn the probabilities
$P(x_1),\ldots,P(x_n)$.

There are less trivial examples where prior information on the
statistical dependencies strongly reduce the required sample size.
For instance, this information may stem from knowledge on the
underlying causal structure. Following \cite{Pearl:00,Spirtes} we
encode causal structure in a directed graph with random variables as
its nodes.  Here we assume the graph to be acyclic. The decisive prior
information assumed to be given here is that each variable has at most
$\Delta$ parents, i.e., is influenced directly by at most $\Delta$ other nodes.
Note that we do not assume that we know which nodes are the parents.
Therefore, our assumption is merely a kind of {\em simplicity
assumption} on the causation for the statistical
dependencies. Furthermore, it should be emphasized that in many cases
one will not find any pair of variables that are statistically
independent. The constraints on the causal structure for the joint
probability measure are more sophisticated and are only reflected in
{\it conditional} probabilities. These constraints are well-known as
the {\it Markov condition} in {\it Bayesian networks} 
\cite{Pearl:00,Glymour}. Conversely,
Bayesian networks may be considered as a convenient and intuitive way
of encoding statistical dependencies among variables in a graph
(without any causal interpretation).

\section{Bayesian networks}
Let us briefly introduce Bayesian networks. To do that we define {\em
conditional independence} relationships among variables, a central
notion in the analysis of probability distributions.

\begin{definition}[Conditional independence]${}$\\ 
Let ${\bf V}=\{X_1,X_2,\ldots,X_n\}$ be a finite set of variables. Let
$P(\cdot)$ be a joint probability distribution over the variables in
$V$, and let ${\bf X}$, ${\bf Y}$ and ${\bf Z}$ stand for any three
subsets of ${\bf V}$. The sets ${\bf X}$ and ${\bf Y}$ are said to be
conditionally independent given ${\bf Z}$, denoted by
\begin{equation}
({\bf X} \perp {\bf Y}\, |\, {\bf Z})
\end{equation} 
if
\begin{equation}
P({\bf x},{\bf y}|{\bf z})=P({\bf x}|{\bf z})P({\bf y}|{\bf z})\,,
\quad\mbox{whenever } P({\bf z})>0\,,
\end{equation}
where ${\bf x}$ is the tuple denoting a particular realization of the
values of the variables in ${\bf X}$ and the tuples ${\bf y}$ and
${\bf z}$ are defined analogously. In words, if all the actual values
of the variables in ${\bf Z}$ are known the actual values of the
variables in ${\bf Y}$ do not provide any further information on the
actual values of the variables in ${\bf X}$.
\end{definition}

Directed acyclic graphs or Bayesian networks -- a term coined in
\cite{Pearl:85} -- are used to facilitate economical representation of
joint probability distributions. The basis decomposition scheme
offered by directed acyclic graphs can be illustrated as follows. Let
$P(\cdot)$ be a joint probability distribution as in Definition~1. The
chain rule of probability calculus always permit to decompose $P$ as
a product of $n$ conditional probability distributions:
\begin{equation}
P(x_1,\ldots,x_n)=\prod_{j=1}^n P(x_j|x_1,\ldots,x_{j-1})\,.
\end{equation}
Now suppose that the conditional probability of some variable $X_j$ is
not sensitive to all the predecessors of $X_j$ but only to a small
subset of those predecessors. In words, suppose that $X_j$ is
independent of all other predecessors, once we know the values of a
selected group of predecessors called ${\bf
P}_j:=\{X_{j,1},\ldots,X_{j,m_j}\}$. We can then write
\begin{equation}
P(x_1,\ldots,x_n)=\prod_{j=1}^n P(x_j|{\bf p}_j)
\end{equation}
considerably simplifying the input information. Instead of specifying
the probability of $X_j$ conditional on all possible realizations of
its predecessors $X_1,\ldots,X_{j-1}$, we need only to take into
account the possible realizations of the set ${\bf P}_j$. The set
${\bf P}_j$ is called the {\em Markovian parents} of $X_j$, or the
parents for short. The reason for the name becomes clear when we
introduce graphs around this concept.

\begin{definition}[Markov parents]${}$\\
Let $V=\{X_1,\ldots,X_n\}$ be an ordered set of variables, and let
$P(\cdot)$ be the joint probability distribution on these
variables. A set of variables ${\bf P}_j$ is said to be Markovian
parents of $X_j$ if ${\bf P}_j$ is a minimal set of predecessors of
$X_j$ that renders $X_j$ independent of all its other predecessors. In
words, ${\bf P}_j$ is any subset of $\{X_1,\ldots,X_{j-1}\}$ satisfying
\begin{equation}\label{eq:markovian}
P(x_j|{\bf p}_j)=P(x_j|x_1,\ldots,x_{j-1})
\end{equation}
such that no proper subset of ${\bf P}_j$ satisfies
Eq.~(\ref{eq:markovian}).
\end{definition} 

This definition assigns to each variable $X_j$ a selected set ${\bf
P}_j$ of preceding variables that are sufficient for determining the
probability of $X_j$. The values of the other preceding variables are
redundant once we know the values ${\bf p}_j$ of the parent set ${\bf
P}_j$. This assignment can be encoded in a directed acyclic graph in
which the variables are represented by the nodes and arrows are drawn
from each node of the parent set toward the child node $X_j$.

Furthermore, Definition~2 also provides a simple recursive method for
constructing such a DAG: Starting with the pair $(X_1,X_2)$, we draw
an arrow from $X_1$ to $X_2$ if and only if the two variables are
dependent. Assume that we have constructed the DAG up to node
$j-1$. At the $j$th stage, we select any minimal set of predecessors
of $X_j$ that renders $X_j$ independent from its other predecessors
(as in Eq.~(\ref{eq:markovian})), call this set ${\bf P}_j$ and draw
an arrow from each member in ${\bf P}_j$ to $X_j$. The result is a
directed acyclic graph, called a Bayesian network, in which an arrow
from $X_i$ to $X_j$ assigns $X_i$ as a Markovian parent of $X_j$,
consistent with Definition~2.

Let us mention that the set ${\bf P}_j$ is unique whenever the
distribution $P(\cdot)$ is strictly positive, i.e.\ every
configuration of variables, no matter how unlikely, has some finite
probability of occurring. Under such conditions, the Bayesian network
associated with $P(\cdot)$ is unique, given the ordering of the
variables \cite{Pearl:88}.

\begin{definition}[Markov Compatibility]${}$\\
Let $G$ be a DAG. If a probability distribution $P$ admits a
factorization relative to $G$, i.e.\
\begin{equation}
P(x_1,\ldots,x_n)=\prod_{j=1}^n P(X_j=x_j|{\bf P}_j={\bf p}_j)\,,
\end{equation}
where ${\bf P}_j$ are the parents of the node $X_j$ defined by the
graph $G$, then we say $G$ and $P$ are compatible, or that $P$ is
Markov relative to $G$.
\end{definition}

The problem of learning a Bayesian network usually treated in the
literature is as follows. Given a {\em training set }
$\{\x^1,\ldots,\x^l\}$, find a network that {\em best matches} the
training set (see e.g.\ \cite{CBL:97,FNP:99}), i.e. to determine a
graph $G$ such that $P$ is Markov relative to $G$.

\section{Networks with bounded in-degree}

To motivate our decisive assumption we would like to note that
scientific reasoning always tries to find a simple explanation
for the data (``Occam's Razor''). We are aware of the fact that 
``simplicity'' is hard to formalize.
However, it seems reasonable to try
to explain data by   {\it simple causal graphs}.
Here we may use the {\it in-degree} of the graph
as criterion for simplicity.
It is defined as the
greatest number of parents that occurs. The intuitive meaning
of in-degree $\Delta$ is that no variable is directly influenced
by more than $\Delta$ others.
For $\Delta \ll n$
we call the graph {\it sparse}.
Clearly, the in-degree is only one of the graph theoretical notions that
may be used to define {\it simplicity} of causal explanations; 
we could use e.g.\ the number of
edges.

Let $G$ be an arbitrary DAG with in-degree $\Delta$. Then every
probability measure that is Markovian relative to $G$ is already
determined by the probabilities of all $(\Delta+1)$-tuples. This
follows directly from the decomposition in Eq.~(\ref{eq:markovian})
since the conditional probabilities $P(x_j|{\bf p}_j)$ are the
quotients of the probabilities $P(x_j,{\bf p}_j)$ and $P({\bf p}_j)$
of sizes at most $\Delta+1$ and $\Delta$, respectively. Consequently,
if $G$ is known we can learn the probability measure $P$ by learning
the probabilities of all $\Delta+1$-tuples.

In contrast, we do not assume that we know the exact structure of $G$
but only that its in-degree at most $\Delta$.  Now the situation is more
complicated. Since we do not know the set of parents for any $X_j$, we
do not know which conditional probabilities have to appear in the
factorization in Eq.~(\ref{eq:markovian}). Therefore, it is not
sufficient to know the probabilities of all tuples of size $\Delta+1$
to reconstruct the structure. We have to know the
probabilities of at least all $(\Delta+2)$-tuples to be able to {\it test}
conditional independencies. The following theorem shows that it is
{\it sufficient} to know the probabilities of all $(2\Delta+1)$ tuples.

\begin{theorem}[Graph structure from correlations]${}$\\\label{Algo}
Let $X_1<X_2<\ldots <X_n$ be an ordering of the variables. Assume that
$P$ is a probability measure that is Markov relative to a directed
acyclic graph (DAG) $G$. Let $G$ be consistent with the ordering,
i.e., the graph $G$ contains no arrow from $X_j$ to $X_i$ for $i<j$.
Let $G$ have in-degree $\Delta$ and assume that the probabilities of
all $(2\Delta+1)$-tuples are known. Then we can find a graph
$\tilde{G}$ (possibly different from $G$) that is Markov relative to
$P$ and has at most in-degree $\Delta$.
\end{theorem}
{\bf Proof:}
We can find the correct graph structure by the following iteration:
Draw an arrow from  $X_1$ to $X_2$ if the two variables are dependent.
Assume we have found the correct structure on $X_1,X_2,\dots,X_{j-1}$.

In order to find a possible minimal set $\cP_j$ of parents of $X_j$ 
we proceed as follows:
Let $m:=\min\{j-1,\Delta\}$.
For each $m$-subset  $K\subseteq V_j:=\{X_1,X_2,\dots,X_{j-1}\}$ 
test whether the following statement is true:

$(X_j \perp L \, |\, K)$ for all sets $L$ (disjoint from $K$) 
that contain at most $m$ 
elements.

If this is true, $K$ contains necessarily a set $\cP'_j$ that can be
taken as Markovian parents of $X_j$.  This can be seen as follows:
Choose $L$ such that $(L\cup K) \supseteq \cP_j$ for an arbitrary
minimal choice of parents of $X_j$.  This is possible since $X_j$ has
at most $m$ parents.  Since $L\cup K$ contains the parents of $X_j$ it
renders $X_j$ independent of its predecessors (see the $d$-separation
criteria in \cite{Pearl:88,Spirtes,Glymour}). Formally we have $(X_j
\perp V_j \,|\, L\cup K)$.  By the contraction rule for conditional
independencies (see \cite{Pearl:88}) the statements $(X_j \perp V_j \,
|\, L\cup K)$ and $(X_j \perp L \,|\, K)$ imply $(X_j \perp V_j \,|\,
K)$. Hence $K$ must contain a set $\cP_j'$ that can be viewed as
Markovian parents of $X_j$.

Now we can test whether a proper subset $K'$  of $K$  satisfies
$(X_j \perp L\,| \,K')$ and obtain a minimal set of parents of $X_j$ by
iterating this procedure. \qed

\section{Learning the probabilities of $k$-tuples}
Now we shall present an upper bound on the required sample size in
order to learn the probabilities of all $k$-tuples with good
reliability. Then we can apply this result to the case
$k:=2\Delta +1$.

Let $P(\cdot)$ be a probability distribution over an (ordered) set of
random variables ${\bf V}=\{X_1,\ldots,X_n\}$ taking on values in
$\Omega_j$ for $j=1,\ldots,n$.

Let $X_{j_{1}},\dots,X_{j_{k}}$ be any $k$-subset of ${\bf V}$. We
would like to have a reliable statement on the probability of the
event
$(x_{j_1},\dots,x_{j_k})\in\Omega_{j_1}\times\cdots\times\Omega_{j_k}$,
i.e.\ the probability
\begin{equation}
P(X_{j_1}=x_{j_1},\ldots,X_{j_k}=x_{j_k})\,.
\end{equation}
The problem to determine the sample size required for estimating
reliably the probability of {\it one specific} event 
is a usual problem of statistics. However,
the problem we encounter in learning Bayesian networks is more
sophisticated: we have to be almost sure that the estimated
probabilities of all $(2\Delta+1)$-tuples are sufficiently close to
the real (unknown) probabilities.

The problem to determine whether and how fast 
the relative frequencies of a large
set of events converge {\it uniformly} to their probabilities is
well-known in statistical learning theory \cite{Vapnik:98}.
Statements on uniform convergence rely on the so-called
Vapnik-Chervonenkis dimension (VC-dimension) of the considered set of events.

\begin{definition}[VC dimension]${}$\\
Let $P$ be an unknown probability measure on a probability space
$\Omega$ and $S$ a set of events, i.e., a set of measurable subsets of
$\Omega$. Define the VC-dimension of $S:=(M_\lambda)$ as the largest
number $h$ such that there exist $h$ points
$\omega_1,\omega_2,\dots,\omega_h \in \Omega$ such that the sets
$M_\lambda \cap \{\omega_1,\dots,\omega_h\}$ run over all $2^h$ subsets of
$\{\omega_1,\dots,\omega_h\}$. 
Intuitively, one can consider the sets $M_\lambda$ as classifiers
and the VC-dimension as the largest number of points that
can be classified in all $2^h$ possible ways.
 The VC-dimension is said to be
infinite if such an $h$-subset can be found for all $h\in\N$.
\end{definition}

A trivial upper bound on the VC-dimension is given by the logarithm to
base $2$ of the number of events (in the case that $S$ is finite).

Finite VC-dimension is known to be sufficient and necessary in order to
have uniform convergence of relative frequencies to their probabilities.
Quantitatively, one has the following theorem:

\begin{theorem}[Uniform convergence]${}$\\\label{Uniform}
Let $f(M)$ be the relative frequency of the number of occurrences of
$M$ after $l$ runs. Let $S$ have VC-dimension $h$. Let $R_{\epsilon}$
be the risk (probability) that $S$ contains at least one set $M$ such
that $|f(M)-P(M)|\geq \epsilon$ for an arbitrary positive
$\epsilon$. Then we have
\begin{equation}
R_{\epsilon} <
4 \exp\left\{\left(
\frac{h(1+\ln(2l/h))}{l}-(\epsilon-1/l)^2 \right)l\right\}\,.
\end{equation}
\end{theorem}
{\bf Proof:} see Theorem 4.4. in \cite{Vapnik:98} \qed

This theorem allows to derive a lower bound on the required sample
size in order to estimate the probability of all $k$-tuples. First we
have to define the set of events and give an upper bound on its
VC-dimension.

Let $\Omega:=\Omega_1\times\cdots\times\Omega_n$
be the probability space.
This means that the $j$th
random variable takes on values from $\Omega_j$ for $j=1,\ldots,n$.
The $k$-tuples are characterized by the positions and values the
corresponding random variables take on. Let ${\bf
j}:=\{j_1,j_2,\dots,j_k\}$ be an arbitrary $k$-subset of
$\{1,\dots,n\}$ and ${\bf x}\in \Omega_{j_1},\dots,\Omega_{j_k}$.

We then denote by $M^{\bf j}_{\bf x}$ the event that the random
variables $X_{j_1},X_{j_2},\dots, X_{j_k}$ take on the values
$x_{j_1},\dots,x_{j_k}$. This event corresponds uniquely to a cylinder
set $C^{\bf j}_{\bf x}\subset \Omega$.

An upper bound on the VC-dimension of the set of those  events that
correspond to cylinder sets $C^{\bf j}_{\bf x}$ is easy to
get.  Let $d$ be the maximal
cardinality of the sets $\Omega_j$. Then, for fixed $k$, there exist
at most
\[
d^k {n \choose k} 
\]
such cylinder sets. The first term gives an upper bound on the
possible combinations of values and the second term the number of
different positions. This number is smaller than $(nd)^k$. 
By taking the logarithm to base $2$ we obtain an upper bound on the 
VC-dimension
\begin{equation}\label{eq:upper}
h\le k\log_2 (nd) 
\end{equation}
Obviously, we can use much better bounds for concrete applications,
e.g.\ given by Stirling's approximation (giving a less intuitive
expression but providing a tighter bound). However, this crude upper
bound is sufficient to study the asymptotic behavior.

Now we will present a lower bound on the VC-dimension in order to get
an idea how tight the upper bound in (\ref{eq:upper}) is.

We construct $l$ $n$-tuples with $l:=\lfloor \log_2 (n-k+1) \rfloor$ 
as follows. For each set $\Omega_j$ we choose two different  values
$x_{j;0}$ and $x_{j;1}$ for $j=1,\ldots,n$. This defines a map 
$\phi$ from the set of binary words of length $n$ into $\Omega$ by
setting
\begin{equation}
\phi: b_1 b_2 \ldots b_n \mapsto
x_{1,b_1} x_{2,b_2} \ldots x_{n,b_n}\,.
\end{equation}

Now we define an $l \times n$ matrix $M$ with entries $0$ and $1$ as
follows: The first $k-1$ columns have only $1$ as entries. The next $2^l$
columns are the binary words of length $l$. The remaining
$(n-k+1 -2^l)$ columns can be chosen arbitrarily.

The rows of $M$ correspond
to $n$-tuples by the map $\phi$. Let $\Y$ be the set of those
$n$-tuples and $S$ be an arbitrary subset of $\Y$. $S$ can uniquely be
characterized by a vector $s$ of length $l$ with entries $0$ and $1$
where the $j$-th entry of $s$ indicates whether the $j$-th $n$-tuple
is an element of $S$ or not.  The matrix $M$ contains a column that
coincides with $s$.  Assume it to be the $i$-th column. Than 
$C^{\bf j}_{\bf x}
\cap \Y $ contains exactly those $n$-tuples that are elements of $S$
provided that $C^{\bf j}_{\bf x}$ is chosen as follows.  Let ${\bf j}$ be
$(1,2,\dots,k-1,i)$ and choose ${\bf x}$ as the $k$-tuple
$(x_{1;1},x_{2;1},\dots,x_{k-1;1},x_{i;1})$.
This shows that the cylinder sets corresponding to $k$-tuples 
are able to classify $\Y$ on all $2^l$ possibilities.
Therefore
$\lfloor\log_2 (n-k+1)\rfloor$ is a lower
bound on the VC-dimension of the cylinder sets. Comparing this bound
with the upper bound in (\ref{eq:upper}), we see that it gives the
correct asymptotic behavior in the $O$-notation 
if $k$ and $d$ are considered as constants.

\begin{theorem}
For $\epsilon >0$ let $R_\epsilon$ be the risk that there is a
cylinder set $C^{\bf j}_{\bf x}$ such that its relative frequency
deviates from its probability by more than $\epsilon$. Than
$R_\epsilon$ can be made smaller than any $\delta>0$ while only
increasing the sample size linearly with $n$.
\end{theorem}
{\bf Proof:}
We choose $l$ such that 
\[
\frac{l}{1+\ln(2l)}\frac{(\epsilon -1/l)^2}{2} \geq 
k \log_2 (nd) \,.
\]
This can asymptotically be achieved by increasing $l$ with $O(n)$,
since $l/(1+\ln(2l))\leq l/(\ln (l))$ and the latter term increases 
less than linearly in $l$.

Using our bound 
\[
h\leq k\log_2 (nd)
\]
we obtain
\[
h \leq \frac{(\epsilon-1/l)^2}{2}\frac{l}{1+\ln(2l)}
\]
and get
\[
\frac{h(1+\ln (2l))}{l} \leq \frac{(\epsilon -1/l)^2}{2}\,.
\]
By elementary calculation, this implies
\[
\frac{h(1+\ln(2l/h))}{l} -(\epsilon -1/l)^2 \leq 
-\frac{(\epsilon -1/l)^2}{2}\,.
\]
Using the bound of Theorem 6 this shows that the risk $R_\epsilon$
can even be made to decrease exponentially in $n$ 
while increasing the sample size $l$ only linearly in $n$.
\qed

Note that the sample size has to be chosen such that the deviation of
the relative frequencies from their probabilities is small compared to
the relative frequencies. Then we have a reasonable criterion to
decide for which sets ${\bf X},{\bf Y},{\bf Z}$ of variables we may
assume ${\bf X}$ and ${\bf Y}$ to be independent given ${\bf Z}$. This
criterion is as follows: Based on the error bound of
Theorem~\ref{Uniform} we compute the relative uncertainty of the
conditional probabilities used in the algorithm in the proof of
Theorem~\ref{Algo}. If the observed statistical dependencies are
greater than the uncertainty we assume the variables to be dependent.

\section{Conclusions}
The sample size to learn the joint probability distribution on $n$
nodes does only increase linearly with $n$ if the underlying causal
structure is assumed to be sufficiently simple.  Here we considered
the case that we know that the (unknown) causal graph has at most
in-degree $\Delta$ and a known time order exists.  Than a graph that
is Markov relative to the unknown probability measure can be found
efficiently if only the probabilities of all $(2\Delta+1)$-tuples are
known. They can be learned with linear sample size.  We have shown
this by finding bounds on the VC-dimension of the corresponding
cylinder sets.  We would like to note that the causal structure can at
least be {\it guessed} if only the probabilities of $(2\Delta
+1)$-tuples are known, since they allow to test a large number of
statistical independencies.

\end{document}